\documentclass[lettersize,journal]{IEEEtran}
\usepackage{amsmath,amsfonts}
\usepackage{algorithmic}
\usepackage{algorithm}
\usepackage{array}
\usepackage[caption=false,font=normalsize,labelfont=sf,textfont=sf]{subfig}
\usepackage{textcomp}
\usepackage{stfloats}
\usepackage{url}
\usepackage{verbatim}
\usepackage{graphicx}
\usepackage{cite}
\usepackage{overpic}
\usepackage{bbm}
\usepackage{bm}
\usepackage{upgreek}
\usepackage{booktabs}
\usepackage{amssymb}
\usepackage{color, xcolor}
\usepackage{hyperref}
\usepackage{graphicx}
\usepackage[normalem]{ulem}
\usepackage{booktabs}
\usepackage{multirow}
\useunder{\uline}{\ul}{}
\begin{document}

\title{Edge-Enabled VIO with Long-Tracked Features for High-Accuracy Low-Altitude IoT Navigation}

\author{Xiaohong~Huang, Cui~Yang, and Miaowen Wen 
\thanks{

The authors are with the School of Electronic and Information Engineering, South China University of Technology, Guangzhou 510640, China (e-mail: eehuangxiaohong@mail.scut.edu.cn; yangcui@scut.edu.cn; eemwwen@scut.edu.cn).}
}


\maketitle
\begin{abstract}
This paper presents a visual-inertial odometry (VIO) method using long-tracked features. Long-tracked features can constrain more visual frames, reducing localization drift. However, they may also lead to accumulated matching errors and drift in feature tracking. Current VIO methods adjust observation weights based on re-projection errors, yet this approach has flaws. Re-projection errors depend on estimated camera poses and map points, so increased errors might come from estimation inaccuracies, not actual feature tracking errors. This can mislead the optimization process and make long-tracked features ineffective for suppressing localization drift. Furthermore, long-tracked features constrain a larger number of frames, which poses a significant challenge to real-time performance of the system. To tackle these issues, we propose an active decoupling mechanism for accumulated errors in long-tracked feature utilization. We introduce a visual reference frame reset strategy to eliminate accumulated tracking errors and a depth prediction strategy to leverage the long-term constraint. To ensure real time preformane, we implement three strategies for efficient system state estimation: a parallel elimination strategy based on predefined elimination order, an inverse-depth elimination simplification strategy, and an elimination skipping strategy. Experiments on various datasets show that our method offers higher positioning accuracy with relatively short consumption time, making it more suitable for edge-enabled low-altitude IoT navigation, where high-accuracy positioning and real-time operation on edge device are required. The code will be published at github\footnote[1]{ \href{https://github.com/xiaohong-huang/FLOW-VIO}{https://github.com/xiaohong-huang/FLOW-VIO}}.

\end{abstract}


\section{Introduction}\label{Introduction}
Low-altitude IoT systems, such as unmanned aerial vehicles (UAVs) and drones, are vital for environmental monitoring, disaster response, and smart city infrastructure. A key challenge in these systems is achieving high-accuracy navigation, which is crucial for the precise operation of autonomous platforms. In the field of edge sensing intelligence, integrated sensing and communication (ISAC) technologies are being advanced by various studies~\cite{isac1,isac2,isac3}. These studies collectively emphasize ISAC's potential to enhance system adaptability and efficiency in complex environments. Visual-inertial odometry (VIO), which focuses on fusing visual and inertial data for high-precision positioning and real-time navigation in low-altitude IoT navigation, complements these areas. Together with other research directions, VIO forms a robust foundation for edge sensing intelligence, driving its application in scenarios such as low-altitude IoT navigation.

Research on tightly coupled VIO methods can be traced back to the multi-state constraint Kalman filter (MSCKF)~\cite{msckf}. Subsequently, many tightly coupled VIO methods have been proposed. OKVIS~\cite{okvis} combines keyframes and bundle adjustment to estimate position by using monocular and stereo cameras. VI-DSO~\cite{vi-dso} extends direct sparse odometry (DSO)~\cite{dso} to visual-inertial odometry by integrating inertial observations with photo-metric errors from high-gradient pixels, achieving high accuracy and improved robustness in low-texture areas. DM-VIO~\cite{dm-vio} is also a strategy developed from DSO~\cite{dso}, which leverages delayed marginalization and poses graph bundle adjustment to enhance accuracy. VINS-Mono~\cite{vins-mono} is a VIO system with loop closure, which uses a single camera and a low-cost IMU for state estimation and performs loop closing by using DBoW2 and 4 degrees of freedom (DoF) pose-graph optimization and map-merging. BASALT~\cite{basalt} is another VIO system with loop closure, which enhances accuracy by incorporating nonlinear factors into bundle adjustment.

VIO focuses on providing high-precision pose estimation by using visual and inertial data, while visual-inertial simultaneous localization and mapping (VI-SLAM) aims to simultaneously estimate pose and build a map of the environment. The main advantage of a SLAM map is that it allows matching and using the short-term, mid-term, and long-term data association for bundle adjustment~\cite{orb-slam3}. The mid-term data association is the major difference between VIO and VI-SLAM. The most representative work of VI-SLAM is ORB-SLAM3~\cite{orb-slam3}. To realize mid-term data association, it uses the long-tracked features to constrain the states of keyframes within the sliding window and their co-visible keyframes. Such a strategy utilizes long-tracked features to overcome localization drift and improve positioning performance.

Leveraging long-tracked features can constrain more visual frames, thereby improving estimation accuracy. Long-tracked features offer unique advantages in visual navigation applications. However, when utilizing them within the framework of edge sensing intelligence for low-altitude IoT navigations, two key challenges arise.

First is the efficiency challenge. Long-tracked features constrain a larger number of frames, which places substantial demands on computational resources and can significantly impact the real-time performance of the system. In edge sensing intelligence applications where resources are often limited, efficiently processing long-tracked features without compromising performance is critical. There is a common practice in solving the Gauss-Newton system within VIO that suggests treating feature states as sparse states and marginalizing them first~\cite{okvis}. However, this conventional method may not always be optimal. For instance, when a long-tracked feature is marginalized, it often results in the generation of a high-dimensional dense matrix. This dense matrix can substantially increase computational complexity and memory usage, making it difficult for the solving strategy to operate in real-time. The increased computational load can be particularly problematic in edge computing environments where resources are constrained, thus highlighting the need for more efficient and adaptive algorithms that can handle the challenges posed by long-tracked features while still meeting the demands of edge sensing intelligence applications.

Another significant challenge is tracking drift. As more visual frames are processed, feature matching errors accumulate, leading to tracking drift~\cite{tracking drift}. To make better use of long-tracked features, solving the feature tracking drift problem is crucial. ORB-SLAM3~\cite{orb-slam3} employs three strategies to mitigate the adverse effects of feature tracking drift. First, during feature matching, it utilizes an image pyramid for multi-scale matching to reduce errors caused by scale variations. Additionally, it dynamically adjusts observation covariance based on the pyramid matching layer to further alleviate feature drift. However, the image pyramid primarily addresses scale-related matching errors and cannot correct matching errors induced by viewpoint changes. Second, ORB-SLAM3 applies loss functions (e.g., Huber loss) to adjust visual residual weights (assigning smaller weights to larger errors), a common practice in VIO and VI-SLAM. Yet this approach faces a fundamental conflict: reprojection errors depend entirely on estimated camera poses and map points, meaning increased errors may reflect estimation inaccuracies rather than actual tracking errors. Such improper weight adjustment misguides optimization and prevents effective suppression of localization drift using long-tracked features. Third, it filters out visual observations exceeding predefined reprojection error thresholds. Similar to loss function-based methods, this filtering relies on estimated errors rather than the actual errors, compromising reliability. While effective for large reprojection errors, its accuracy degrades significantly for small errors (where estimation errors may mask tracking errors). Moreover, this strategy shortens the lifespan of long-tracked features and diminishes their impact.

\begin{figure*}[!t]
\centering
\begin{overpic} [width=6.7in]{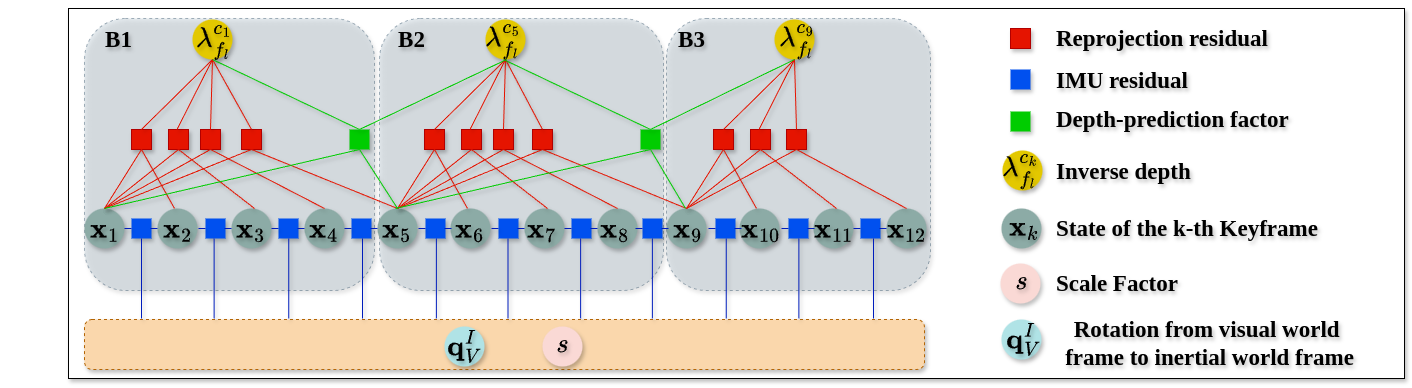}
\end{overpic}
\caption{Illustration of the factor graph of the SWF in a case of the system having 12 keyframes, which are divided into 3 blocks, denoted as $\mathbf{B}_1$, $\mathbf{B}_2$ and $\mathbf{B}_3$. Here, the size of the block is 4. For every long-tracked feature, each time its tracking crosses a block, we change the reference frame, generate a new inverse depth, and predict it using the previous inverse depth.
}\label{fig:systemoverview}
\end{figure*}

Although some works~\cite{tracking drift,pyramid,affine} have attempted to improve the matching accuracy of feature points to reduce tracking drift, these efforts can only decrease the drift under certain conditions and cannot eliminate it entirely. Ultimately, we all inevitably face the challenge of solving a system that involves feature points with tracking drift. In this paper, we focus on how to solve such a system, rather than on how to improve tracking. Most existing methods utilize the long-term constraint characteristics of long-tracked feature points by minimizing the re-projection errors of all observations, which is undoubtedly optimal when the observation errors of feature points obey a zero-mean normal distribution. However, if this assumption is still insisted upon for feature points with tracking drift, erroneous visual observations will undoubtedly lead to incorrect estimation results. This paper propose a novel method for leveraging the long-term constraint characteristics of long-tracked feature points via depth prediction rather than minimizing the re-projection errors of all observations. The key contributions of this work are as follows:
\begin{itemize}
\item[$\bullet$]  To address the issue of inaccurate state estimation caused by tracking drift in long-term feature tracking, we design a new frame structure that divides keyframes within the sliding window filter (SWF)  into multiple blocks, setting the first frame of each block as the reference frame. This allows us to reset the accumulation of re-projection errors by changing the reference frame for a feature point in each block. To leverage the long-term constraint characteristics of long-tracked features, we further propose a depth prediction mechanism that predicts the inverse depth in the reference frame based on the inverse depth in the previous reference frame. 
\item[$\bullet$] To ensure real-time performance, we propose three strategies for efficient system state estimation: a parallel elimination strategy based on predefined elimination order, an inverse-depth elimination simplification strategy, and a elimination skipping strategy. With these three strategies, our system can still operate in real time even with a large-capacity sliding window, where the large-capacity sliding window helps prevent nonlinear observations of long-tracked features from being prematurely linearized.
\end{itemize}

\section{Sliding Window Filter}\label{Sliding Window Filter}

The factor graph of our system with one long-tracked feature is depicted in Fig.~\ref{fig:systemoverview}. In our system, we divide the frames into multiple blocks. If a feature has observations in any two non-adjacent blocks, it is regarded as long-tracked. Otherwise, it is regarded as short-tracked. For a concise presentation, we do not show the factor graph with short-tracked features. The factor graph with short-tracked features can be easily obtained from Fig.~\ref{fig:systemoverview}. For example, removing the prediction factors from Fig.~\ref{fig:systemoverview}, we can obtain a factor graph with 3 short-tracked features. Fig.~\ref{fig:systemoverview} only shows a simple case that the system has 12 frames. In our system with 100 keyframes, we have 10 blocks, and the size of the block is 10. 

In our method, every new coming frame will be put into the SWF for state optimization. If the new frame is not a keyframe, we throw the frame after optimization; otherwise, the frame is retained in the SWF. When the number of keyframes exceeds the capacity of the SWF, we update the SWF prior $\{\mathbf {H}_{\mathcal P},\mathbf {b}_{\mathcal P},\hat{\mathbf {X}}_{\mathcal P}\}$ by marginalizing the oldest block. For instance, in Fig.~\ref{fig:systemoverview}, if the oldest block (i.e., $\mathbf{B}_1$) is marginalized, we would generate a prior to constrain $\mathbf{X}_{\mathcal{P}}=[\mathbf{x}_5,\lambda_{f_l}^{c_5},\mathbf{q}^I_V,s]$. The readers may refer to~\cite{okvis} for details of marginalizing states from measurements to generate a prior.  

\subsection{Notations and Definitions}\label{Notation And Definitions}

We consider $(.)^{V} $ as the visual world frame, $(.)^{I} $ the inertial world frame, $(.)^b$ the IMU frame, $(.)^c$ the camera frame, $(.)^f$ the feature point, and $s$ the scale factor. We use the rotation matrix $\mathbf R$ and Hamilton quaternion $\mathbf q$ to represent rotation. $\mathbf q^V_{c}$ and $\mathbf p^V_{c}$ respectively represent the rotation and translation from the camera frame to the visual world frame.  $\mathbf q^I_{V}$ represents the rotation from the visual world frame to the inertial world frame. $\lambda^{c_j}_{f_l}$ represents the inverse depth of the $l$-th feature, whose reference frame is the $j$-th keyframe in the SWF.

Next, we introduce the reference frame index mapping function ${\hbar(.)}$. Consider $f_l$ as a certain feature, and $\mathbf{\breve{u}}^{c_k}_{f_l}$ as its observation in the $k$-th frame. We define the reference frame index of $\mathbf{\breve{u}}^{c_k}_{f_l}$ as ${\hbar(k,l)}$. If $f_l$ is short tracked, the reference frame of $\mathbf{\breve{u}}^{c_k}_{f_l}$ is always its first observed frame. If $f_l$ is long-tracked,  the reference frame index is set as $\hbar(k,l)=\lfloor (k-2)/M\rfloor\times M+1 $, where $M$ is the size of the block, $\lfloor.\rfloor$ is a round down function. If the long-tracked feature has no observation in the selected reference frame, we change the reference frame to $\hbar(k,l)=\lfloor (k-2)/M\rfloor\times M+1+M $. That is, the reference frame of the long-tracked observation is only set in the first frame of a certain block, which can be used for solving the Gauss-Newton system efficiently (see Section~\ref{Gauss-Newton Solving Strategy}).

\subsection{Problem Formulation}\label{Problem Formulation}
Let $\mathbf q^V_{c_{k}}$, and $\mathbf p^V_{c_{k}}$ be the orientation and position of the camera frame under visual world frame, respectively;  $\mathbf v^I_{b_{k}}$ be the velocity of the IMU frame under inertial world frame. Denote the IMU bias by $\mathbf b_{k}$. Let $\mathbf x_k=[\mathbf p^V_{c_{k}},\mathbf q^V_{c_{k}},\mathbf v^I_{b_{k}},\mathbf b_{k}]$ be a state vector of the $k$-th frame. Then the state of the system, denoted by $\mathbf X$, can be represented as: 
\begin{align}
\mathbf X=\{s,\mathbf{q}^I_V\}\cup \bigcup_{k\in \mathcal{S}_{\text{SWF}} } \mathbf{x}_k\cup \bigcup_{l,j\in \mathcal{S}_{\lambda} } \lambda^{c_j}_{f_l},
\label{eq:1}
\end{align}
where $\mathcal{S}_{\text{SWF}}$ is a set of frames remained in the SWF, and $\mathcal{S}_{\lambda}$ is a set of inverse depths in the SWF.

We use the Gauss-Newton method to obtain a maximum posterior estimation by minimizing the sum of the Mahalanobis norm of all residuals:
\begin{equation}
\begin{split}
\hat {\mathbf X}=\mathop{\arg\min}_{\mathbf X} \{&\bm{r}_{\mathcal{P}}^T\left(\mathbf{X}_{{\mathcal P}}\right)\mathbf{H}_{{\mathcal P}}^{-1}\bm{r}_{\mathcal{P}}\left(\mathbf{X}_{{\mathcal P}}\right)
\\&+\sum _{k\in \mathcal {S_I}}\bm{r}_{\mathcal{I}_k}^T\left(\mathbf{X}\right)\mathbf {P}_{\mathcal{I}_k}^{-1}\bm{r}_{\mathcal{I}_k}\left(\mathbf{X}\right)
\\&+\sum _{\left(k,l\right)\in \mathcal {S_V}}\bm{r}_{\mathcal{V}_{k,l}}^T\left(\mathbf{X}\right)\mathbf{P}_{\mathcal{V}_{k,l}}^{-1}\bm{r}_{\mathcal{V}_{k,l}}\left(\mathbf{X}\right)
\\&+\sum _{\left(k,l\right)\in \mathcal {S_D}}\bm{r}_{\mathcal{D}_{k,l}}^T\left(\mathbf{X}\right)\mathbf{P}_{\mathcal{D}_{k,l}}^{-1}\bm{r}_{\mathcal{D}_{k,l}}\left(\mathbf{X}\right)
\},
\end{split}
\label{eq:2} 
\end{equation}
where $\bm{r}_{\mathcal{I}_k}$ and $\bm{r}_{\mathcal{V}_{k,l}}$ are residuals for IMU and visual measurements, respectively; $\mathbf {P}_{\mathcal{I}_k}$ and $\mathbf{P}_{\mathcal{V}_{k,l}}$ are the corresponding covariances; $\mathcal {S_I}$ and $\mathcal {S_V}$ are the sets of IMU and visual measurements, respectively; $\bm{r}_{\mathcal{D}_{k,l}}$ is the prediction residual of the inverse depth; $\mathbf{P}_{\mathcal{D}_{k,l}}=[\sigma_d^2]_{1\times 1}$ is the prediction covariance; $\mathcal {S_D}$ is the set of inverse depth of long-tracked features; $\bm{r}_{\mathcal{P}}\left(\mathbf{X}_{{\mathcal P}}\right)=\mathbf{H}_{{\mathcal P}}\delta \mathbf{X}_{{\mathcal P}}-\mathbf{b}_{{\mathcal P}}$ is the residual of the prior; $\delta \mathbf{X}_{{\mathcal P}}$ is the incremental vectors of $ \mathbf{X}_{{\mathcal P}}$ relative to $ \hat{\mathbf{X}}_{{\mathcal P}}$; $\mathbf{H}_{\mathcal P}^{-1}$ is the pseudo inverse~\cite{okvis} of $\mathbf{H}_{\mathcal P}$.

Let $\mathbf{p}^I_{b_k}=\mathbf{R}^I_V(s\mathbf{p}^V_{c_k}+\mathbf{R}^V_{c_k}\mathbf{p}^c_b)$ and $\mathbf{R}^I_{b_k}=\mathbf{R}^I_V\mathbf{R}^V_{c_k}\mathbf{R}^c_b$ be the position and orientation of the IMU frame under inertial world frame, respectively, where $\{\mathbf{p}^c_b,\mathbf{R}^c_b\}$ is the IMU-camera extrinsic parameters. The IMU residual $\bm{r}_{\mathcal{I}_k}(\mathbf{X})$ is defined  in~\cite{vins-mono}.

The residual for the $l$-th feature observation is defined as:
\begin{align}
&\bm{r}_{\mathcal{V}_{k,l}}(\mathbf{X})=\bm{\pi}(\mathbf p^{c_k}_{f_l})-\mathbf{\breve{u}}^{c_k}_{f_l},
\label{eq:3} 
\end{align}
with
\begin{align}
&\mathbf p^{c_k}_{f_l}=\mathbf R_{c_{k}}^{V^T}(\mathbf p^{V}_{f_l}-\mathbf p^V_{c_k}),
\label{eq:4} 
\\&
\mathbf p^{V}_{f_l}=\frac{1}{\lambda^{c_j}_{f_l}}\mathbf R_{c_{j}}^{V}\bm{\pi}^{-1}(\mathbf{\breve{u}}^{c_j}_{f_l})+\mathbf p_{c_{j}}^{V},\ j={\hbar(k,l)}
\label{eq:4_2} 
\end{align}
where ${\hbar(.)}$ is a reference frame mapping function, and $\bm{\pi}(.)$ is the perspective projection function. $\mathbf{\breve{u}}^{c_k}_{f_l}$ and $\mathbf{\breve{u}}^{c_j}_{f_l}$ are the measurement of the $l$-th feature at frames $k$ and $j$, respectively. 

The prediction residual of the inverse depth is defined as:
\begin{align}
&\bm{r}_{\mathcal{D}_{k,l}}(\mathbf{X})=\frac{1}{[\mathbf p^{c_k}_{f_l}]_z}-\lambda^{c_k}_{f_l},
\label{eq:5} 
\end{align}
where $\mathbf p^{c_k}_{f_l}$ is defined in~(\ref{eq:4}) and (\ref{eq:4_2}). $[.]_z$ extracts the $z$ component of a positional vector. The prediction covariance can be set to small value to tightly constrain the new inverse depths. The readers may use the public solving library (e.g., the Ceres-Solver~\cite{ceres}) for solving the Gauss-Newton system of (\ref{eq:2}) by setting the prediction covariance as: $\sigma_d=10^{-5}$). In our system, we use a new solving strategy to solve the Gauss-Newton system, and the prediction covariance is regarded as 0. We will detail our solving strategy in Section~\ref{Simplification of the Inverse depth Prediction}. 

\begin{figure}[!t]
\centering
\begin{overpic} [width=3.5in]{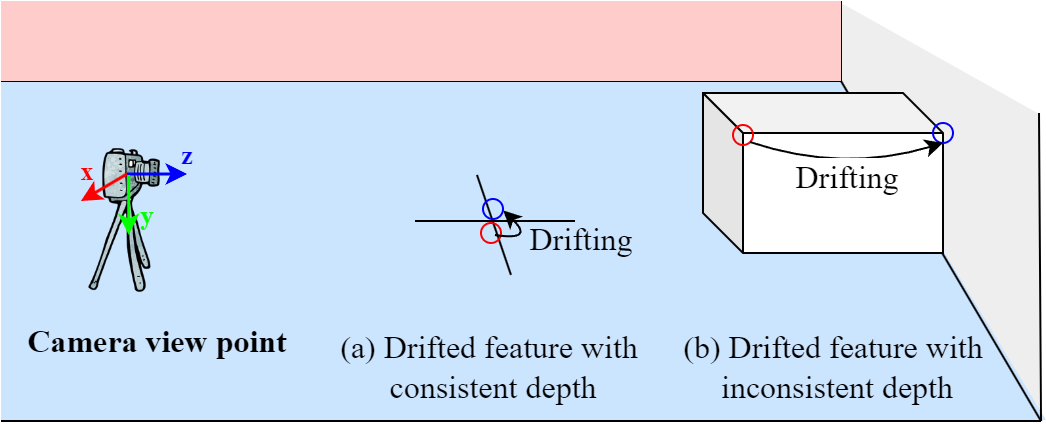}
\end{overpic}
\caption{Illustration of the feature tracking drifts with consistent and inconsistent depth. The red and blue circles represent the positions of the first observed feature and drifted features, respectively. Consider $\lambda_1$ is the inverse depth of the first-observed feature, $\lambda_2$ is the inverse depth of the drifted feature. In (a), we have $|\frac{1}{\lambda_1}-\frac{1}{\lambda_2}|\ll\frac{1}{\lambda_i},\forall\lambda_i\in\{\lambda_1,\lambda_2\}$. Therefore, we consider the drifted feature in (a) has consistent depth, which often occurs when the feature points fall on a flat surface. In (b), the drifted feature has inconsistent depth since the condition mentioned above is not satisfied, which often occurs when the feature points fall on sharp edges.
}\label{fig:drifting}
\end{figure}

\subsection{Depth Drift Detection}\label{Depth Drift Detection}
In our system, we assume that the predicted inverse depth is equal to the actual inverse depth of the observation in the new reference frame. However, such an assumption may not always be correct. The tracking drift may also generate a depth drift, making the predicted inverse depth differ greatly from the actual inverse depth of the new observation (see Fig.~\ref{fig:drifting}.b). We need a strategy to detect the depth drift and reject the measurements that cause depth drift.

The depth drift will cause re-projection errors if there are translation excitation between the re-projection frames and the reference frame. Therefore, we detect the depth drift using the re-projection errors in multiple frames. More specifically, for every inverse depth  $\lambda^{c_j}_{f_l}$, we use ($\ref{eq:4_2}$) to compute the feature position $\mathbf{p}_{f_l}^V$, and use $\mathbf{p}_{f_l}^V$ to compute the re-projection errors in its following $K$ frames $\{c_{j+1},c_{j+2},..., c_{j+K}\}$. If the averaged re-projection error is larger than a certain threshold, $s_a$, or if a single re-projection error is larger than another  threshold, $s_b$, we consider the depth drift has occurred, and remove the corresponding measurements in those $K$ frames.

Different values of $K$ will lead to different results. If $K$ is too small, there may have no enough translation excitation for detecting the depth drift. If $K$ is too large, the tracking drift may accumulate to a significant value, and hence make it not appropriate for detection. In Section~\ref{Experimental Results}, we will present the results for different values of $K$.

\section{Gauss-Newton Solving Strategy}\label{Gauss-Newton Solving Strategy}
The multi-reference frames  and depth-prediction strategies not only address the tracking drift problem but also enable us to design a new method to solve the system efficiently. There is a common sense that the feature states should be regarded as sparse states, and marginalize them first when solving the system. The purpose to do so is to remain the sparsity of the system.  However, marginalizing the feature states first may not always be a good option. For example, when a long-tracked feature is being marginalized, it will generate a dense matrix with high dimensions, which will make the solving strategy difficult to operate in real-time.  In our method, not all the feature states are marginalized first.
\subsection{Parallel Elimination Strategy}
\begin{figure}[!t]
\centering
\begin{overpic} [width=3.5in]{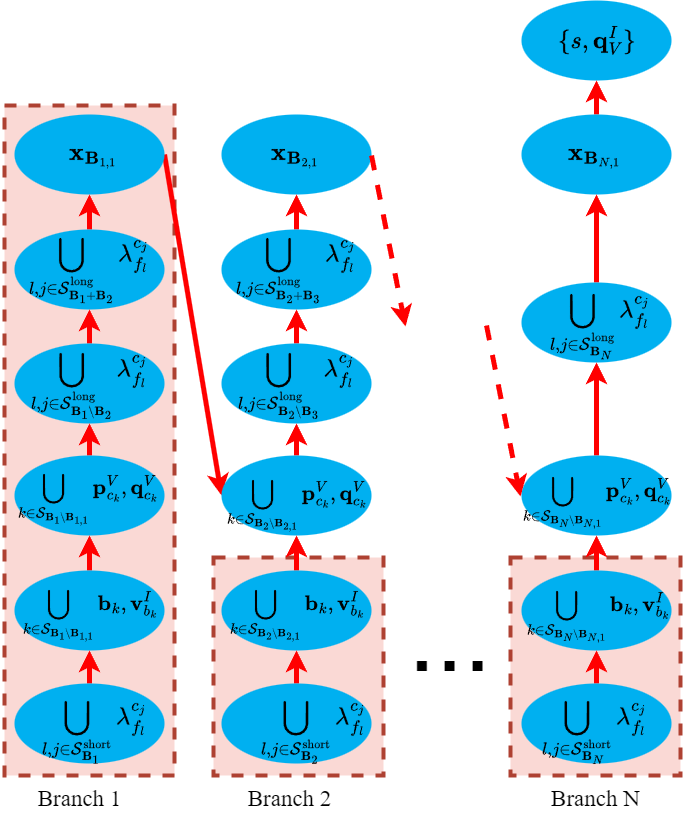}
\end{overpic}
\caption{Illustration of the elimination tree. Here, $\mathcal{S}_{\mathbf{B}_i\backslash{\mathbf{B}_{i,1}}}$ is the set of frame indexes in the block $\mathbf{B}_i$,  excluding its first frame index; $\mathcal{S}_{\mathbf{B}_i}^{\text{short}}$ is the set of inverse depth of the short-tracked feature that has observations in block $\mathbf{B}_i$; $\mathcal{S}_{\mathbf{B}_i+\mathbf{B}_{i+1}}^{\text{long}}$ is the set of inverse depth that has observation in block $\mathbf{B}_i$ and block $\mathbf{B}_{i+1}$; $\mathcal{S}_{\mathbf{B}_i\backslash \mathbf{B}_{i+1}}^{\text{long}}$ is the set of inverse depth that has observation in block $\mathbf{B}_i$ but not in $\mathbf{B}_{i+1}$; $\mathbf{x}_{\mathbf{B}_{i,j}}$ is the state vector of the $j$-th frame in the $i$-th block. Blue ellipses represent the super-nodes. Red arrows represent the elimination order. Pink boxes represent the independent branches. The elimination of the independent branches can be implemented in parallel.
}\label{fig:elimination_tree}
\end{figure}

In this section, we introduce our developed strategy for solving the Gauss-Newton problem.
There are multiple strategies~\cite{sparse_1}\cite{sparse_2} that make use of the sparsity  to speed up the solving process by factorizing the hessian matrix as $\mathbf{H=LL^T}$ and solving the system by $\mathbf{LL^T\triangle X=b}$. The major consumption time is the factorization of $\mathbf{H}$, which is largely subject to the elimination order applied~\cite{etree}. Our solving strategy is similar to the process described in~\cite{sparse_1}, except that we use a predefined elimination order rather than the approximate minimum degree (AMD) order~\cite{AMD}, and our elimination process is implemented using the Schur-Complement strategy rather than the Cholesky factorization.  An elimination tree is given in Fig.~\ref{fig:elimination_tree}   to illustrate our predefined elimination order and the independent branches, where the elimination for the independent branches can be implemented in parallel.

In our previous study~\cite{ours}, we have proposed a parallel elimination strategy with another elimination order for solving the Gauss-Newton problem. Here, we briefly introduce how to solve the Gauss-Newton system given a certain elimination tree.  The readers may refer to~\cite{gauss-schur} for more details of using the elimination tree and Schur-Complement to solve a Gauss-Newton system. Consider that $\mathcal{A}$ is a certain set of states that are selected to be eliminated. $\mathbf{H}\triangle\mathbf{X}=\mathbf{b}$ is a Gauss-Newton system constructed by the measurements that have connections with $\mathcal{A}$. $\mathcal{B}$ is a set of states that have connections with $\mathcal{A}$. Hence the reduced Gauss-Newton system after $\mathcal{A}$ is being eliminated can be defined as $\mathbf{H}^*\triangle\mathbf{X}_{\mathcal{B}}=\mathbf{b}^*$, with
\begin{align}
&\mathbf{H}^*=\mathbf{H}_{\mathcal{B}\mathcal{B}}-\mathbf{H}_{\mathcal{B}\mathcal{A}}\mathbf{H}_{\mathcal{A}\mathcal{A}}^{-1}\mathbf{H}_{\mathcal{A}\mathcal{B}},\label{eq:8_1} 
\\&\mathbf{b}^*=\mathbf{b}_{\mathcal{B}}-\mathbf{H}_{\mathcal{B}\mathcal{A}}\mathbf{H}_{\mathcal{A}\mathcal{A}}^{\mathbf{-1}}\mathbf{b}_{\mathcal{A}},\label{eq:8_2} 
\\&\mathbf{H}=\begin{bmatrix}
\mathbf{H}_{\mathcal{A}\mathcal{A}}&\mathbf{H}_{\mathcal{A}\mathcal{B}}
\\\mathbf{H}_{\mathcal{B}\mathcal{A}}&\mathbf{H}_{\mathcal{B}\mathcal{B}}\end{bmatrix},
\mathbf{b}=\begin{bmatrix}
\mathbf{b}_{\mathcal{A}}
\\\mathbf{b}_{\mathcal{B}}
\end{bmatrix}.\label{eq:8_3} 
\end{align}

The reduced Gauss-Newton system is delivered to the next super-node and generates a new Gauss-Newton system, and eliminates the new set of selected states to generate a new reduced Gauss-Newton system, and so on until only one super-node is retained. We solve the last Gauss-Newton system to generate the incremental estimation of the last super-node and perform a backward substitution~\cite{schur} to compute all the incremental estimations.

There are some reasons why we use the predefined order rather than the AMD order~\cite{AMD}. First, the consumption time for AMD ordering is not cheap (about equal to the consumption time for performing single factorization). Although our predefined order will have a slightly higher fill-in~\cite{AMD}, we find that the consumption time induced by the extra fill-in is much smaller than the consumption time used for AMD ordering. Second, we have found a novel strategy based on our elimination order for eliminating $\bigcup_{l,j\in \mathcal{S}^{\text{long}}_{\mathbf{B}_i + \mathbf{B}_{i+1}}}\lambda^{c_j}_{f_l}$, which will be detailed in Section~\ref{Simplification of the Inverse depth Prediction}. Third, based on our elimination order, we use a new strategy to skip the elimination of a certain set of old blocks, which will be described in Section~\ref{Elimination Skipping Strategy}. Fourth, with our elimination order, we can easily obtain the prior when performing the marginalization strategy, which is the reduced Gauss-Newton system after $\mathbf{x}_{\mathbf{B}_{1,1}}$ is eliminated.

\subsection{Simplification of the Inverse Depth Prediction}\label{Simplification of the Inverse depth Prediction}
Consider  $\mathcal{A}_i$ is a subset of states of block $\mathbf{B}_i$, where $\mathcal{A}_i$ is defined as:
\begin{align}
& \mathcal{A}_i=\bigcup_{l,j\in \mathcal{S}^{\text{short}}_{\mathbf{B}_i}}\lambda^{c_j}_{f_l}\cup 
\bigcup_{k\in \mathcal{S}_{\mathbf{B}_i\backslash{\mathbf{B}_{i,1}}} } \mathbf{x}_{k}\cup 
\bigcup_{l,j\in \mathcal{S}^{\text{long}}_{\mathbf{B}_i \backslash \mathbf{B}_{i+1}}}\lambda^{c_j}_{f_l}.
\end{align}
 From Fig.~\ref{fig:systemoverview} and~\ref{fig:elimination_tree}, the set of states that have connections with $\mathcal{A}_i$ can be defined as: $\mathcal{B}_i=\mathcal{C}_i\cup
\mathcal{D}_i$, where $\mathcal{C}_i$ and $\mathcal{D}_i$ are respectively defined as:
\begin{align}
&\mathcal{C}_i=\{\mathbf{x}_{\mathbf{B}_{i,1}},\mathbf{x}_{\mathbf{B}_{i+1,1}},s,\mathbf{q}^I_V\}\cup\bigcup_{k\in\mathcal{S}_{\mathbf{B}_{i+1}}}\{\mathbf{p}^V_{c_k},\mathbf{q}^V_{c_k}\}
, \\& \mathcal{D}_i=\bigcup_{l,j\in \mathcal{S}^{\text{long}}_{\mathbf{B}_i + 
\mathbf{B}_{i+1}}}\lambda^{c_j}_{f_l}.\label{eq:7} 
\end{align}
The reduced Gauss-Newton system after $\mathcal{A}_i$ is eliminated can be defined as: $\mathbf{H}^*_i\triangle\mathbf{X}_{\mathcal{B}_i}=\mathbf{b}^*_i$. We can eliminate $\mathcal{D}_i$ similar to the prediction procedure of the Kalman filter. From~(\ref{eq:5}), the new inverse depth after prediction can be defined as:
\begin{align}
&\lambda^{c_k}_{f_l}=\frac{1}{[\mathbf p^{c_k}_{f_l}]_z}+\eta_d,\ \eta_d\sim\mathcal{N}(0,\sigma_d^2),
\label{eq:55} 
\end{align}
where $\mathbf p^{c_k}_{f_l}$ is defined in~(\ref{eq:4}) and (\ref{eq:4_2}). $\eta_d$ is the prediction noise. Consider $\mathbf{H}^+_i\triangle\mathbf{X}_{\mathcal{B}_i^+}=\mathbf{b}^+_i$ is a new reduced Gauss-Newton system after $\mathcal{D}_i$ is eliminated, by using the prediction procedure of Kalman filter to eliminate $\mathcal{D}_i$, $\mathbf{H}^+_i$ and $\mathbf{b}^+_i$ can be respectively defined as:
\begin{align}
&\mathbf{H}^+_i=(\mathbf{J}_i{\mathbf{H}^*_i}^{-1}\mathbf{J}_i^{\mathbf{T}}+\mathbf{Q}_i)^{-1},
\mathbf{b}_i^+=\mathbf{H}^+_i\mathbf{J}_i{\mathbf{H}^*_i}^{-1}\mathbf{b}^*_i,\label{eq:9} 
\end{align}
with
\begin{align}
\mathbf{J}_i=\frac{\partial \mathbf{X}_{\mathcal{B}_i^+}}{\partial \mathbf{X}_{\mathcal{B}_i}^T},\label{eq:10} 
\end{align}
where $\mathcal{B}_i^+=\mathcal{C}_i\cup\mathcal{F}_i$, with $\mathcal{F}_i$ being the set of the new inverse depth after prediction. $\mathbf{Q}_i=\text{diag}(\sigma_d^2,...,\sigma_d^2)$ is the prediction covariance matrix. The Jacobians of $\mathbf{X}_{\mathcal{C}_i}$ with respect to $\mathbf{X}_{\mathcal{C}_i}$ and $\mathbf{X}_{\mathcal{D}_i}$ are $\mathbf{I}$ and $\mathbf{0}$, respectively, and the Jacobian of $\mathbf{X}_{\mathcal{F}_i}$ with respect to $\mathbf{X}_{\mathcal{B}_i}$ can be delivered from~(\ref{eq:55}). In our system, we consider the prediction covariance as 0, and $\mathbf{H}^+_i$ and $\mathbf{b}^+_i$ can be simplified as:
\begin{align}
&\mathbf{H}^+_i=\mathbf{J}_i^{\mathbf{-T}}\mathbf{H}^*_i\mathbf{J}_i^{\mathbf{-1}},\mathbf{b}_i^+=\mathbf{J}_i^{\mathbf{-T}}\mathbf{b}^*_i.\label{eq:99} 
\end{align}
Note that $\mathbf{J}_i$ is a sparse matrix, whose inverse $\mathbf{J}_i^{\mathbf{-1}}$, which has the same adjacency matrix as $\mathbf{J}_i$, is easy to be computed. $\mathbf{J}_i^{\mathbf{-1}}$ is also sparse, which makes the multiplication with $\mathbf{J}_i^{\mathbf{-1}}$ easy to be computed.  In Section~\ref{Notation And Definitions}, we define the reference frames of the long-tracked features as only set in the first frame of its observed blocks. The reason to do so is that it can reduce the dimension of $\mathbf{J}_i$.

\begin{table}[]
\centering
\caption{The average number of features tracked within the SWF of our proposed, categorized by different tracking ranges intervals.}
\label{tab:tab0}
\resizebox{\columnwidth}{!}{%
\begin{tabular}{ccccccc}
\hline
\multicolumn{1}{c|}{} & \multicolumn{1}{c|}{} & $[2,10)$ & $[10,30)$ & $[30,50)$ & $[50,90)$ & $[90,100)$ \\ \hline
\multicolumn{1}{c|}{\multirow{3}{*}{\begin{tabular}[c]{@{}c@{}}EuRoC\\ Dataset\end{tabular}}} & \multicolumn{1}{c|}{MH} & 1130 & 269.3 & 67.5 & 34.4 & 5.6 \\
\multicolumn{1}{c|}{} & \multicolumn{1}{c|}{V1} & 1448 & 293.1 & 59.1 & 24.2 & 1.7 \\
\multicolumn{1}{c|}{} & \multicolumn{1}{c|}{V2} & 1397 & 260.4 & 55.7 & 23.4 & 2 \\ \hline
 &  &  &  &  &  &  \\ \hline
\multicolumn{1}{c|}{\multirow{5}{*}{\begin{tabular}[c]{@{}c@{}}TUM\\ Dataset\end{tabular}}} & \multicolumn{1}{c|}{corridor} & 949 & 155.3 & 19.8 & 6.9 & 2.6 \\
\multicolumn{1}{c|}{} & \multicolumn{1}{c|}{magistrale} & 1083 & 146.8 & 26.7 & 16.1 & 10.4 \\
\multicolumn{1}{c|}{} & \multicolumn{1}{c|}{outdoors} & 940 & 132 & 25.8 & 15.8 & 14.3 \\
\multicolumn{1}{c|}{} & \multicolumn{1}{c|}{room} & 867 & 178.9 & 28.1 & 10.5 & 4 \\
\multicolumn{1}{c|}{} & \multicolumn{1}{c|}{slides} & 994 & 134.6 & 26.9 & 17.5 & 12.4 \\ \hline
 &  &  &  &  &  &  \\ \hline
\multicolumn{1}{c|}{\multirow{2}{*}{\begin{tabular}[c]{@{}c@{}}Our \\ Dataset\end{tabular}}} & \multicolumn{1}{c|}{M1} & 740 & 208.5 & 52.7 & 32.2 & 23.9 \\
\multicolumn{1}{c|}{} & \multicolumn{1}{c|}{M2} & 1278 & 418.6 & 26.9 & 4.3 & 0.3 \\ \hline
 &  &  &  &  &  &  \\ \hline
\multicolumn{1}{c|}{\multirow{7}{*}{\begin{tabular}[c]{@{}c@{}}4Seasons\\ Dataset\end{tabular}}} & \multicolumn{1}{c|}{office} & 2711 & 189 & 17 & 4.5 & 0.3 \\
\multicolumn{1}{c|}{} & \multicolumn{1}{c|}{neighbor} & 2290 & 222.8 & 25.4 & 7.7 & 0.6 \\
\multicolumn{1}{c|}{} & \multicolumn{1}{c|}{business} & 2351 & 215.6 & 23.9 & 6.1 & 0.4 \\
\multicolumn{1}{c|}{} & \multicolumn{1}{c|}{country} & 1875 & 191.7 & 22 & 9.6 & 2.2 \\
\multicolumn{1}{c|}{} & \multicolumn{1}{c|}{city} & 2563 & 163.7 & 10.6 & 3.1 & 0.3 \\
\multicolumn{1}{c|}{} & \multicolumn{1}{c|}{oldtown} & 2229 & 162.8 & 16.8 & 6.8 & 2.3 \\
\multicolumn{1}{c|}{} & \multicolumn{1}{c|}{parking} & 1294 & 198.7 & 38.4 & 16.3 & 0.2 \\ \hline
\end{tabular}%
}
\end{table}

\subsection{Elimination Skipping Strategy}\label{Elimination Skipping Strategy}
We use the nonlinear-cost-change to describe whether the linearization points of a certain block are properly set. Consider $n_i$ is the nonlinear-cost-change of block $\mathbf{B}_i$, where $n_i$ is defined as:
\begin{align}
n_i=\|\bm r_{\mathbf{B}_i}(\mathbf{\hat X}_0)+\mathbf{J}_{\mathbf{B}_i}(\mathbf{\hat X}_0)(\mathbf{\hat X}-\mathbf{\hat X}_0)-\bm r_{\mathbf{B}_i}(\mathbf{\hat X})\|^2,\label{eq:11} 
\end{align}
where $\mathbf{\hat X}_0$ is the linearization point vector, $\mathbf{\hat X}$ is the current estimation, and $\bm r_{\mathbf{B}_i}(.)$ and $\mathbf{J}_{\mathbf{B}_i}(.)$ are the residual vector and the corresponding Jacobian matrix of block $\mathbf{B}_i$, respectively. If $n_i$ is larger than a certain threshold (e.g., $10^{-3}$ in our system), we update the Jacobian matrix, the residual vector, and the corresponding Gauss-Newton system of the block and its following blocks $\mathbf{B}_i,\mathbf{B}_{i+1},...,\mathbf{B}_N$. Otherwise, we skip the updation and the corresponding matrix elimination (detailed in (\ref{eq:8_1})) of the block $\mathbf{B}_i$, and only remain the updation of its right-hand side vector $\mathbf{b}_i=\mathbf{J}_{\mathbf{B}_i}(\mathbf{\hat X}_0)^{\mathbf{T}}\bm r_{\mathbf{B}_i}(\mathbf{\hat X})$ and the vector elimination (detailed in (\ref{eq:8_2})) of the new $\mathbf{b}_i$.

In general, the older frame will have smaller value of $n_i$, which is also the reason why we set the older frame in the head of the elimination order.

\section{Experimental Results}\label{Experimental Results}

Our method runs on a computer with an AMD Core 5800 3.2 GHz $\times$ 8 core CPU using the Robot Operating System (ROS). Our frontend is built on VINS-Mono~\cite{vins-mono} and uses the KLT sparse optical flow algorithm~\cite{optical-flow} for feature tracking. Compared with descriptor-based matching, optical flow achieves a higher frame-to-frame matching success rate and generally lower short-term tracking error, making it better suited for our framework. In our system, the number of blocks and block size are both set to 10, resulting in an SWF capacity of 100.  We extract about 200 features per frame. The thresholds used for detecting the depth drift are set to $[s_a,s_b]=[4\sigma_v,12\sigma_v]$, where $\sigma_v$ is the standard deviation of the visual measurements. We evaluate our method on the EuRoC dataset~\cite{EuRoC}, the TUM-VI dataset~\cite{tum}, the 4Seasons dataset~\cite{4 seasons} and our dataset~\cite{ours}. The EuRoC dataset is recorded by a drone, while the TUM-VI and our dataset are captured with a handheld device, and the 4Seasons dataset is recorded with a vehicle. The frame number used for detecting depth drift is set to $K=50$ for the EuRoC, TUM-VI, and our datasets.  Because the travel velocity of the camera in the 4Seasons dataset is very high (about 10 times faster than other datasets), a smaller value of $K$ can generate significant translation excitation for detecting depth drift. Therefore, we set $K=20$ for the 4Seasons dataset. Table~\ref{tab:tab0} shows the average number of feature points tracked within the SWF of our proposed  system, categorized by different tracking range intervals. We use the absolute trajectory error (ATE) or localization drift error (i.e., drift=$\frac{\text{ATE}\cdot 100}{\text{length}}$) to show the positioning errors. All methods are evaluated 10 times for EuRoC and 5 times for the other datasets on each sequence.

\begin{table}[]
\caption{Comparison of the ATE [m] and the average consumption time [ms] between various VIO and SLAM systems on the EuRoC dataset. Best results in bold, underline is the second-best result.}
\label{tab:1}
\resizebox{\columnwidth}{!}{%
\begin{tabular}{c|ccccc|c}
\hline
 & \multicolumn{5}{c|}{VIO} & SLAM \\ \hline
 & VINS & \begin{tabular}[c]{@{}c@{}}Open\\ VINS\end{tabular} & BASALT & DM-VIO & Ours & \begin{tabular}[c]{@{}c@{}}ORB-\\ SLAM3\end{tabular} \\
 & mono & stereo & stereo & mono & mono & mono \\ \hline
MH1 & 0.15 & 0.072 & 0.07 & 0.065 & \textbf{0.046} & {\ul 0.052} \\
MH2 & 0.15 & 0.143 & 0.06 & {\ul 0.044} & \textbf{0.029} & 0.079 \\
MH3 & 0.22 & 0.086 & {\ul 0.07} & 0.097 & 0.084 & \textbf{0.055} \\
MH4 & 0.32 & 0.173 & 0.13 & {\ul 0.102} & \textbf{0.053} & 0.114 \\
MH5 & 0.30 & 0.247 & 0.11 & \textbf{0.096} & {\ul 0.106} & {\ul 0.106} \\
V11 & 0.079 & 0.055 & {\ul 0.04} & 0.048 & 0.049 & \textbf{0.038} \\
V12 & 0.11 & 0.06 & 0.05 & {\ul 0.045} & 0.053 & \textbf{0.023} \\
V13 & 0.18 & 0.059 & 0.10 & 0.069 & {\ul 0.054} & \textbf{0.048} \\
V21 & 0.08 & 0.054 & 0.04 & {\ul 0.029} & \textbf{0.028} & 0.053 \\
V22 & 0.16 & 0.047 & 0.05 & 0.050 & {\ul 0.037} & \textbf{0.029} \\
V23 & 0.27 & - & 0.22 & 0.114 & {\ul 0.071} & \textbf{0.051} \\
avg & 0.184 & 0.096 & 0.085 & 0.069 & \textbf{0.055} & {\ul 0.059} \\ \hline
\begin{tabular}[c]{@{}c@{}}consumption\\ time {[}ms{]}\end{tabular} & 16.7 & 5.12 & 11.5 & 46.9 & 22.2 & 116 \\ \hline
\end{tabular}%
}
\end{table}

\begin{table}[]
\caption{Comparison of the ATE [m] between various VIO and SLAM systems on  the TUM-VI dataset.}
\label{tab:2}
\resizebox{\columnwidth}{!}{%
\begin{tabular}{c|ccccc|c}
\hline
 & \multicolumn{5}{c|}{VIO} & SLAM \\ \hline
 & ROVIO & VINS & BASALT & DM-VIO & \multicolumn{1}{c|}{Ours} & \begin{tabular}[c]{@{}c@{}}ORB-\\ SLAM3\end{tabular} \\
 & stereo & mono & stereo & mono & \multicolumn{1}{c|}{mono} & mono \\ \hline
corridor1 & 0.47 & 0.63 & 0.34 & {\ul 0.19} & 0.37 & \textbf{0.08} \\
corridor2 & 0.75 & 0.95 & 0.42 & 0.47 & {\ul 0.16} & \textbf{0.02} \\
corridor3 & 0.85 & 1.56 & 0.35 & {\ul 0.24} & 0.26 & \textbf{0.02} \\
corridor4 & {\ul 0.13} & 0.25 & 0.21 & {\ul 0.13} & \textbf{0.07} & 0.15 \\
corridor5 & 2.09 & 0.77 & 0.37 & {\ul 0.16} & 0.31 & \textbf{0.02} \\
magistrale1 & 4.52 & 2.19 & \textbf{1.20} & 2.35 & 1.73 & {\ul 1.64} \\
magistrale2 & 13.43 & 3.11 & 1.11 & 2.24 & \textbf{0.77} & {\ul 0.79} \\
magistrale3 & 14.80 & \textbf{0.40} & 0.74 & 1.69 & {\ul 0.50} & 3.96 \\
magistrale4 & 39.73 & 5.12 & 1.58 & 1.02 & {\ul 0.59} & \textbf{0.17} \\
magistrale5 & 3.47 & 0.85 & {\ul 0.60} & 0.73 & \textbf{0.52} & 1.21 \\
magistrale6 & - & 2.29 & 3.23 & {\ul 1.19} & \textbf{0.75} & 1.76 \\
outdoors1 & 101.95 & {\ul 74.96} & 255.04 & 123.24 & \textbf{42.38} & 111 \\
outdoors2 & 21.67 & 133.46 & 64.61 & {\ul 12.76} & \textbf{3.59} & 19.6 \\
outdoors3 & 26.10 & 36.99 & 38.26 & {\ul 8.92} & \textbf{2.16} & \textbf{-} \\
outdoors4 & - & 16.46 & 17.53 & {\ul 15.25} & \textbf{2.3} & 26.5 \\
outdoors5 & 54.32 & 130.63 & 7.89 & {\ul 7.16} & \textbf{1.92} & 16.9 \\
outdoors6 & 149.14 & 133.60 & 65.50 & {\ul 34.86} & \textbf{17.16} & 39.1 \\
outdoors7 & 49.01 & 21.90 & {\ul 4.07} & 5.00 & \textbf{2.87} & 11.0 \\
outdoors8 & 36.03 & 83.36 & 13.53 & \textbf{2.11} & {\ul 2.56} & 29.8 \\
room1 & 0.16 & 0.07 & 0.09 & \textbf{0.03} & {\ul 0.04} & \textbf{0.03} \\
room2 & 0.33 & 0.07 & 0.07 & 0.13 & {\ul 0.06} & \textbf{0.01} \\
room3 & 0.15 & 0.11 & 0.13 & 0.09 & \textbf{0.04} & {\ul 0.07} \\
room4 & 0.09 & 0.04 & 0.05 & 0.04 & {\ul 0.03} & \textbf{0.02} \\
room5 & 0.12 & 0.20 & 0.13 & 0.06 & {\ul 0.04} & \textbf{0.02} \\
room6 & 0.05 & 0.08 & {\ul 0.02} & {\ul 0.02} & 0.03 & \textbf{0.01} \\
slides1 & 13.73 & 0.68 & 0.32 & {\ul 0.31} & \textbf{0.21} & 0.93 \\
slides2 & 0.81 & 0.84 & {\ul 0.32} & 0.87 & \textbf{0.17} & 0.87 \\
slides3 & 4.68 & 0.69 & 0.89 & {\ul 0.60} & \textbf{0.52} & 1.12 \\
avg drift \% & 16.83 & 1.70 & 0.94 & {\ul 0.47} & \multicolumn{1}{c|}{\textbf{0.19}} & 0.67 \\ \hline
\end{tabular}%

}
\end{table}

\begin{figure}[!t]
\centering
\begin{overpic} [width=3.5in]{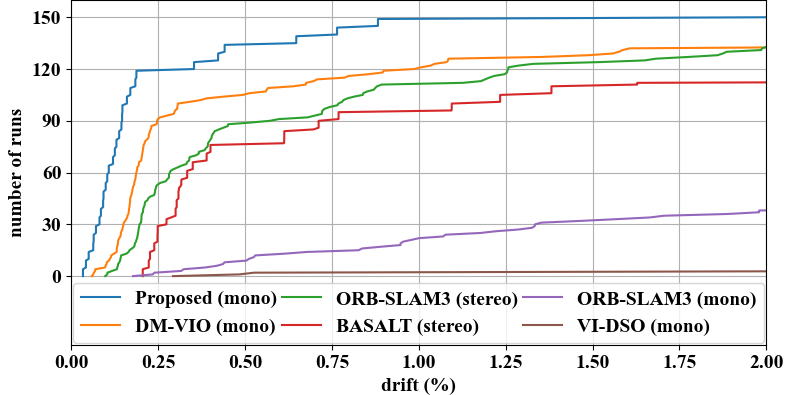}
\end{overpic}
\caption{ Cumulative error plot for the 4Seasons dataset (drift in $\%$).
}\label{fig:4seasons}
\end{figure}

We compare the results with VINS-Mono~\cite{vins-mono}, OpenVINS~\cite{openvins}, BASALT~\cite{basalt}, DM-VIO~\cite{dm-vio}, ROVIO~\cite{rovio}, VI-DSO~\cite{vi-dso}, and ORB-SLAM3~\cite{orb-slam3}. For a fair comparison, we disable explicit loop closure detection and re-localization for all the methods. Note that ORB-SLAM3 can still achieve local loop closure through its co-visibility graph mechanism after explicit loop closure detection is turned off. Table~\ref{tab:1} shows the comparison in the EuRoC dataset, and Table~\ref{tab:2} shows the comparison in the TUM-VI dataset. Following~\cite{dm-vio}, results of the 4Seasons dataset are presented in cumulative error plot, which show how many sequences (y-axis) have been tracked with an accuracy better than the threshold on the x-axis. As shown in Tables~\ref{tab:1} and \ref{tab:2} and Fig.~\ref{fig:4seasons}, by comparing the results between different VIO systems, we can see that almost all the best and second-best results are generated by our method. 
The reason that our system has better performance than other VIO systems is described below. When our system only has short-tracked features, its framework is similar to that of other small SWF methods (i.e., all the VIO methods mentioned above). Like other small SWF methods, short-tracked features can provide high-precision short-term relative pose estimation by constructing re-projection constraints from common or consecutive blocks. The long-tracked features of our system can reduce the localization drift, thereby improving accuracy. Consider  $\hat{\lambda}_{f_l}^{c_j}$ as the inverse depth estimation of a feature in its first observed block, and  $\hat{\lambda}_{f_l}^{c_{j+N\cdot M}}$ as the inverse depth prediction in its  $N$-th subsequent block, which is computed by a sequential prediction using (\ref{eq:55}): $\hat{\lambda}_{f_l}^{c_j}\to\hat{\lambda}_{f_l}^{c_{j+M}}...\to \hat{\lambda}_{f_l}^{c_{j+N\cdot M}}$. From (\ref{eq:55}), we can see that as the localization (translation or rotation) drift error increases, the error in $\hat{\lambda}_{f_l}^{c_{j+N\cdot M}}$ also increases, leading to larger re-projection errors (see (\ref{eq:3})). By minimizing the re-projection errors, the localization drift error can be reduced.

\begin{figure}[!t]
\centering
\begin{overpic} [width=3.5in]{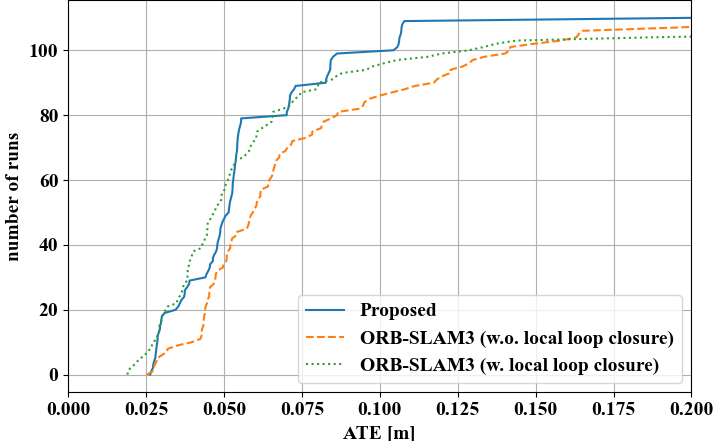}
\end{overpic}
\caption{The cumulative ATE plot of our proposed and ORB-SLAM3 on the EuRoC dataset.
}\label{fig:orb}
\end{figure}

By comparing with the results of ORB-SLAM3, it can be observed that ORB-SLAM3 only achieves satisfactory performance on datasets with small-scale motions (such as the EuRoC datasets, the corridor types, and the room types of the TUM-VI datasets). This is because ORB-SLAM3 additionally incorporates local loop closure functionality~\cite{dso} compared to other methods (including the proposed method). ORB-SLAM3 can detect local loops through its covisibility graph mechanism. When the camera moves repeatedly within a small scene, ORB-SLAM3 can frequently and implicitly close numerous small loops along with some large ones. However, in large-scale motion scenarios (e.g., the 4Seasons datasets, the magistrate, the outdoors, and the slide types of the TUM-VI datasets), the probability of loop occurrence decreases substantially, leading to degraded performance of ORB-SLAM3. Figure~\ref{fig:orb} further demonstrates the localization results of ORB-SLAM3 on the EuRoC dataset after disabling its local loop closure function using~\cite{dso}. As shown in Fig.~\ref{fig:orb}, after disabling the local loop closure function of ORB-SLAM3, its performance declines significantly and falls below that of our approach. Moreover, even enableing the local loop closure function of ORB-SLAM3, the proposed method demonstrates comparable localization accuracy and superior robustness to ORB-SLAM3. Although loop closure correction can partially eliminate accumulated drift, it essentially serves as an opportunistic correction mechanism. While loop closures may frequently occur and provide corrections for indoor applications like service robots or cleaning robots, their occurrence probability drastically reduces in outdoor large-scale scenarios such as autonomous driving, logistics vehicles, and cruise drones. In such cases, the system localization accuracy primarily depends on the precision of incremental motion estimation. Although ORB-SLAM3 also utilizes long-track features, it only considers tracking drift caused by image scale variations. For other scenarios, it uses the reprojection error to adjust the fusion weight, which diminishes the advantages of long-tracked features.

Table~\ref{tab:1} further displays the average computation time for processing a single frame by different algorithms. For ORB-SLAM3, we only include the consumption times for its LocalMapping thread to process one frame.  Note that the computational time for feature extraction and matching is not included, as these processes can be implemented in parallel with the processing of the SWF. As shown in Table~\ref{tab:1}, although our system does not have the least computation time, the computation time of our system is not significantly higher compared to other methods. 

\begin{table}[]
\centering
\caption{Comparison of the ATE [m] between various VIO and GNSS systems on  real world experiment.}
\label{tab:3}
\begin{tabular}{c|cc|cc|c}
\hline
 & \multicolumn{2}{c|}{VIO} & \multicolumn{2}{c|}{GNSS-VIO} & GNSS-only \\ \hline
 & VINS & Ours & RVINS & GVINS & \multirow{2}{*}{RTKLIB} \\
 & mono & mono & mono & mono &  \\ \hline
R1M1 & 1.55 & \textbf{0.37} & \textbf{0.37} & {\ul 0.78} & 2.31 \\
R2M1 & 1.43 & \textbf{0.38} & {\ul 0.67} & 1.02 & 4.34 \\
R1M2 & 1.55 & 1.32 & \textbf{0.43} & {\ul 0.56} & 1.43 \\
R2M2 & 2.64 & \textbf{1.27} & 1.43 & {\ul 1.31} & 4.20 \\
avg & 1.79 & {\ul 0.84} & \textbf{0.73} & 0.92 & 3.07 \\ \hline
\end{tabular}%
\end{table}

\begin{figure}[!t]
\centering
\begin{overpic} [width=3.3in]{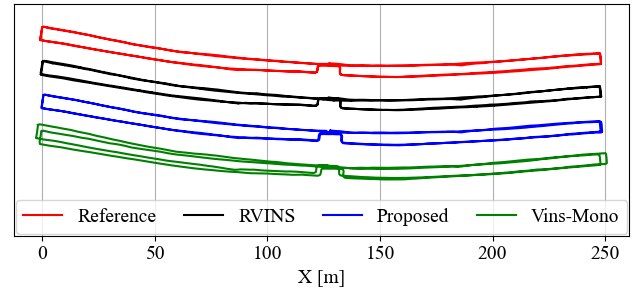}
\end{overpic}
\caption{Positioning results of RVINS, VINS-Mono and ours in R1M1.
}\label{fig:multiple_comparision}
\end{figure}
\begin{figure*}[!t]
\centering
\begin{overpic} [width=7in]{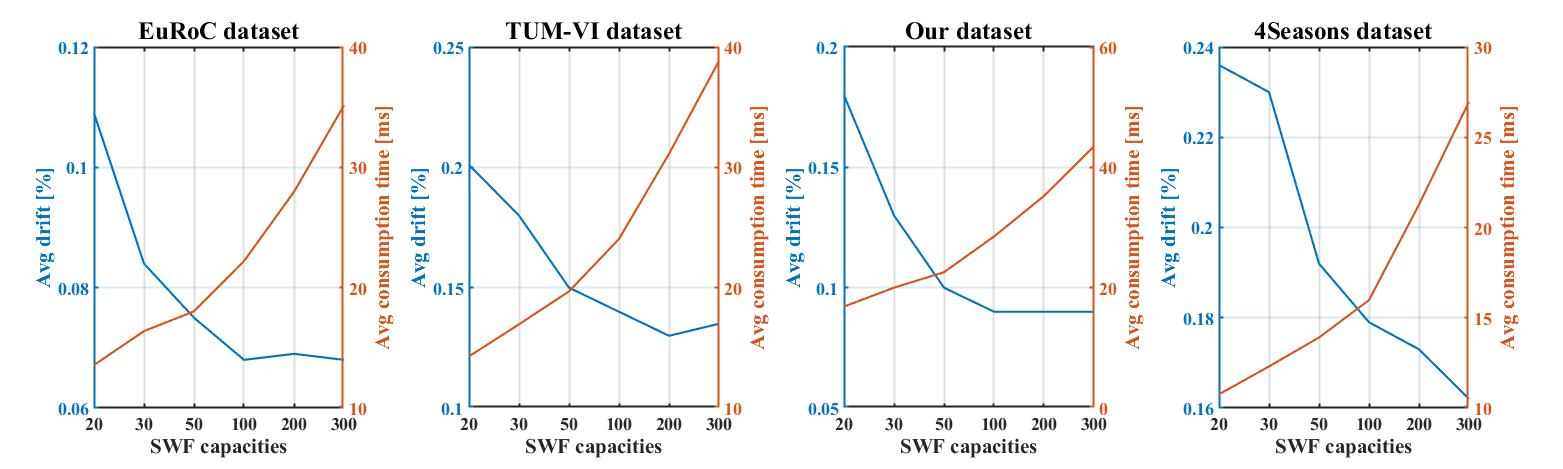}
\end{overpic}
\caption{Illustration of the average drift error and average consumption time of the system with different SWF capacities. Only the average consumption times for the SWF processing one visual frame are shown in the plot. The consumption times for feature extraction are not included.}\label{fig:different_capacity}
\end{figure*}
\begin{table}[]
\centering
\caption{Illustration of the number of outliers in the results of our system under different SWF capacities.}
\label{tab:4}
\begin{tabular}{lllllll}
\hline
SWF capacity & 20 & 30 & 50 & 100 & 200 & 300 \\ \hline
TUM-VI dataset & 6 & 4 & 2 & 2 & 2 & 2 \\ \hline
4Seasons dataset & 7 & 7 & 4 & 3 & 3 & 2 \\ \hline
\end{tabular}
\end{table}
\begin{figure}[!t]
\centering
\begin{overpic} [width=3.2in]{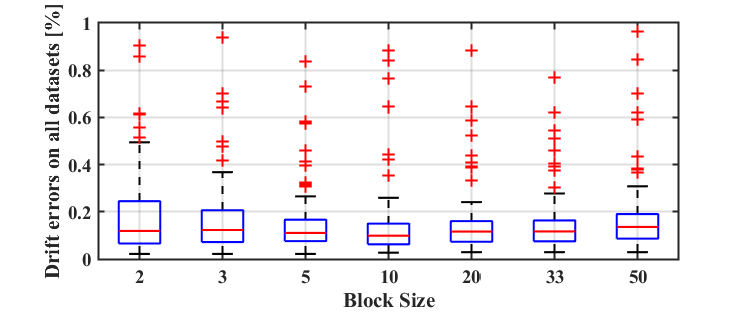}
\end{overpic}
\caption{Drift errors of the system for different values of  block sizes.
}\label{fig:different_sizein}
\end{figure}
\begin{figure}[!t]
\centering
\begin{overpic} [width=3.2in]{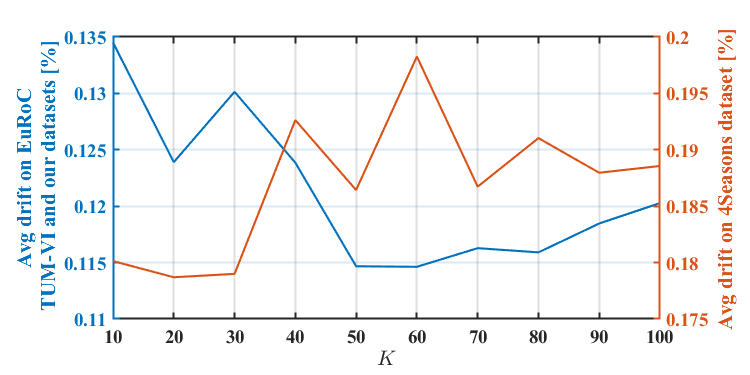}
\end{overpic}
\caption{Average drift of the system for different values of  $K$.
}\label{fig:differentK}
\end{figure}

In the TUM-VI and 4Seasons datasets, we demonstrated that our system generates accurate estimations in outdoor environments. Here, we further compare our system with various GNSS fusion systems to analyze its performance over short travel distances (e.g., a few kilometers).  We evaluate our method in our dataset~\cite{ours} and compare the results with different GNSS-VIO systems. Our dataset includes experiments in two regions, R1 and R2, where R1 is GNSS-friendly and R2 is GNSS-challenged. For each region, we conduct two experiments, labeled M1 and M2. In M1, the camera moves steadily with low velocity and angular rate, while in M2, we rotate the yaw angle during movement. The travel distance is about 1 km for one experiment. We compare the results with VINS-Mono~\cite{vins-mono}, RVINS~\cite{ours}, GVINS~\cite{gvins} and RTKLIB~\cite{rtklib}. RVINS and GVINS are two GNSS-VIO fusion systems. RTKLIB~\cite{rtklib} is a GNSS-only system. The results are illustrated in Table~\ref{tab:3}.  As shown in Table~\ref{tab:3}, our system outperforms others in the GNSS-challenged region (R2M1 and R2M2). In the GNSS-friendly region, when the camera moves stably (R1M1), our system also matches the performance of GNSS-VIO systems. As an example, Fig.~\ref{fig:multiple_comparision} shows the positioning results of RVINS, VINS-Mono and our proposed in R1M1. The trajectory of R1M1 is combined by two circles. As shown in Fig.~\ref{fig:multiple_comparision}, the two circles estimated by our system are almost overlapping. However, for VINS-Mono, there is a large gap between the two circles.

Next, we evaluate our method with different SWF capacities. We use the average drift to measure the overall performance. However, we found that some results have very large drift errors, which would affect the reliability of the average drift. Therefore, results with drift errors greater than 0.5\% are considered outliers, and their drift errors are capped at 0.5\%. Fig.~\ref{fig:different_capacity} shows the average drift and average consumption time of the system when the SWF capacity is set to 20, 30, 50, 100, 200, and 300, respectively. Table~\ref{tab:4} shows the corresponding outlier number. Note that datasets not appearing in Table~\ref{tab:4} indicate that there are no outliers for those datasets. Here, the size of the block is 10, and we adjust the number of blocks to change the capacity of the SWF. As shown in  Fig.~\ref{fig:different_capacity} and Table~\ref{tab:4},  as the capacity of the SWF increases, the positioning accuracy improves, and the number of outliers decreases accordingly. We can also see that the consumption time also increases with the increase of the capacity. However, the increased quantity is acceptable since we use the parallel elimination strategy and the elimination skipping strategy. 

We evaluate the performance of the system using different block sizes. Fig.~\ref{fig:different_sizein} shows the results when the block sizes are set to $M=[2,3,5,10,20,33,50]$. Note that when adjusting the block size, we also modify the number of blocks to  $\lfloor100/M\rfloor$, ensuring that the SWF capacity is approximately equal to 100.  As shown in Fig.~\ref{fig:different_sizein}, both large and small block sizes can degrade system performance. The reasons are described as follows. If the block size is too small, each blocks lack sufficient translation excitation, leading to low-accuracy depth estimation. In contrast to most other methods, our approach first estimates the depth of features within each block and then fuses them using depth prediction factors to obtain depth estimation in the first observed block, followed by predictions for other blocks. If the translation excitation in the blocks are all insufficient, even if the final combined excitation is large, our method cannot achieve high-accuracy depth estimation, resulting in decreased system performance. Conversely, if the block size is too large, tracking drift increases significantly, further lowering system performance. Note that $M=10$ may not always be the best option in our system. For example, if we set the average parallax~\cite{vins-mono} from 10 pixels to 5 pixels, the best $M$ in Fig.~\ref{fig:different_sizein} would be changed to around 25.

We further evaluate our method with different values of $K$. Fig.~\ref{fig:differentK} shows the average positioning drift of the system with $K$ set to different values. As mentioned above, if $K$ is too small, there may be no enough translation excitation for detecting the depth drift. If $K$ is too large, the tracking drift may accumulate to a significant value, and hence make it inappropriate for detection. Therefore, as shown in Fig.~\ref{fig:differentK}, both large and small $K$ values can deteriorate the performance of the system on the EuRoC, TUM-VI and our datasets. Because the travel velocity in 4Seasons dataset is very high, only the large $K$ value reduce the performance of the system.

\section{Conclusion and Future Work}
This paper presents a visual-inertial odometry framework tailored for low-altitude IoT navigation, resolving the inherent tension between long-tracked feature utilization and edge computing constraints. While long-tracked features enhance localization accuracy by constraining multiple frames, they introduce tracking drift from accumulated matching errors and computational complexity incompatible with real-time edge deployment. Our solution integrates a multi-block reference frame reset strategy to decouple drift propagation and a depth prediction mechanism to maintain geometric consistency, effectively suppressing erroneous feature correlations. To address computational demands, we propose parallel elimination, inverse-depth simplification, and elimination skipping strategies, collectively optimizing the Gauss-Newton solving process for edge devices. Experimental validation confirms that our method offers higher positioning accuracy with relatively short consumption time. Moreover, experimental results showed that our method is able to achieve better positioning performance, especially with large-scale movement, which makes it more appropriate for outdoor environments. This work will be combined with our previous work~\cite{ours} to develop a more accurate RTK-visual-inertial system in the future.

\end{document}